%% file: Template.tex
\documentclass{article}
\usepackage{spconf}
\input{inputfiles/packages}

\title{On-the-fly Text Retrieval for End-to-End ASR Adaptation}
\name{Bolaji Yusuf$^{1,2,3}$\sthanks{Work done as an Applied Scientist intern at Amazon Alexa.}, Aditya Gourav$^1$, Ankur Gandhe$^1$, Ivan Bulyko$^1$}
\address{
$^{1}$ Amazon Alexa, USA \\
$^{2}$ Boğaziçi University, Department of Electrical and Electronics Engineering, Istanbul, Turkey \\
$^{3}$ Brno University of Technology, Faculty of Information Technology, Speech@FIT, Czechia}
\begin{document}
\ninept
\maketitle
\begin{abstract}
\input{inputfiles/abstract}
\end{abstract}
\section{Introduction}
\label{sec:intro}
\input{inputfiles/reintro}

\section{Model}
\label{sec:model}
\input{inputfiles/remodel}

\section{Experiments}
\label{sec:experiments}
\input{inputfiles/reexperiments}

\section{Conclusions}
\label{sec:conclusions}
\input{inputfiles/conclusion}

\clearpage
\bibliographystyle{IEEEbib}
\bibliography{refs}

\appendix
\newpage
\section{Ablations}
\label{sec:extras}
\input{inputfiles/extras}

\end{document}

%% file: inputfiles/packages.tex
\usepackage{amsmath,graphicx}
\usepackage{amssymb}
\usepackage{bbm}
\usepackage{textcomp}
\usepackage{mathtools}
\usepackage{booktabs}
\usepackage[utf8]{inputenc}
\usepackage[dvipsnames]{xcolor}
\usepackage{mystyle}
\usepackage{multirow}
\usepackage{tikz}
\usepackage{hyperref}
\usepackage{pgfplots}
\usepackage{tikz-3dplot}
\usepackage{pifont}
\usepackage{subcaption}
\usepackage{numprint} 
\usepackage{diagbox}
\usepackage{cite}
\usetikzlibrary{arrows.meta}
\usetikzlibrary{positioning,through, backgrounds,decorations.pathreplacing,shapes.misc,patterns}
\pgfplotsset{compat=1.17}

\tdplotsetmaincoords{70}{30}

\usepackage{nccmath}
\usepackage{xpatch}
\xpatchcmd{\NCC@ignorepar}{%
\abovedisplayskip\abovedisplayshortskip}
{%
\abovedisplayskip\abovedisplayshortskip%
\belowdisplayskip\belowdisplayshortskip}
{}{}

%% file: inputfiles/abstract.tex
End-to-end speech recognition models are improved by incorporating external text sources, typically by fusion with an external language model.
Such language models have to be retrained whenever the corpus of interest changes.
Furthermore, since they store the entire corpus in their parameters, rare words can be challenging to recall.
In this work, we propose augmenting a transducer-based ASR model with a retrieval language model, which directly retrieves from an external text corpus plausible completions for a partial ASR hypothesis.
These completions are then integrated into subsequent predictions by an adapter, which is trained \emph{once}, so that the corpus of interest can be switched without incurring the computational overhead of retraining.
Our experiments show that the proposed model significantly improves the performance of a transducer baseline on a pair of question-answering datasets. Further, it outperforms shallow fusion on recognition of named entities by about $7\%$ relative; when the two are combined, the relative improvement increases to $13\%$.

%% file: inputfiles/reintro.tex
End-to-end (E2E) speech recognition models can be improved on a domain when they are shown text from that domain.
While there have been works that do so by training parts of the model on external text~\cite{wang21t_interspeech,yusuf22usted,thomas2022integrating}, the most common method of incorporating text, unless precluded by computational constraints, is still fusion with language models (LM)~\cite{gulcehre2015using,Chorowski2017,variani2020hybrid,mcdermott2019density,meng2021internal} since they can be swapped at inference time.

Nevertheless, even LMs, especially neural LMs, can be unwieldy to change to match user interest.
It is common for users of voice assistants and other speech technologies to use words and phrases associated with trending topics such as sporting events, album releases, pandemics etc.
To contend with these surges in user interest, these ASR systems must be able to rapidly assimilate words of interest and also gracefully discard them as such ephemeral interest wanes.
Although it is possible to obtain relevant text from internet forums or news articles, incorporating them into the ASR LM requires retraining the entire LM or training a separate LM for each trending topic.
Moreover, LMs struggle with proper nouns and other named entities which are of the most interest because such entities, by nature rare and diffuse in training data, are assigned low likelihoods by LMs which store all information in their parameters.

This has sparked interest in biasing methods which attempt to boost the likelihoods of a catalog of entities.
These include finite state transducer (FST)-based methods which compose the ASR output with an FST with negative-cost arcs carrying the entities to be boosted~\cite{zhao2019shallow,gourav2021personalization,kocour2021interspeech,pundak2018deep}, and deep-biasing methods which introduce a trainable adapter into the ASR model, with an attention mechanism to select the right entity to boost~\cite{pundak2018deep,jain20_interspeech,feng2021,sathyendra2022contextual,dingliwal2022}.
However, both are more suited to catalogs of limited size (up to a few hundred at a time), such as contact names and song playlists, rather than the large catalogs necessary to cover multiple trending topics at the same time.
FSTs for instance boost all items in the catalog with predefined weights making it hard to control what gets boosted as the catalog size increases.
For deep-biasing, the limitation is due to the smearing of attention weights as the catalog size increases, as well as the increasing computational overhead of multihead attention.
Therefore, it remains a challenge to have a rapidly adaptable, computationally efficient way to bias ASR towards large lists of phrases--possibly millions of tokens--at a time.

Inspired by the success of retrieval mechanisms in language modeling~\cite{khandelwal2020generalization,guu2020retrieval,he2021efficient}, we propose  augmenting an RNN transducer (RNN-T)-based ASR model with a retriever which searches in an external datastore for candidate continuations of a partial ASR hypothesis.
The RNN-T's encoder output then attends to encodings of the retrieved continuations, and the attention output is summed to the encoder output before begin fed into the joiner.

Experiments on the Squad~\cite{rajpurkar-etal-2018-know} and DeepMind Question-Answering~\cite{hermann2015teaching} datasets show that a strong RNN-T baseline can be improved by retrieving from datastores that contains related text, even when the datastores also have millions of tokens of unrelated, distracting content.
Furthermore, retrieval can be complemented by shallow fusion as the latter performs better on recognition of common words while retrieval performs better for named entities.

%% file: inputfiles/remodel.tex
\subsection{RNN Transducer}
\begin{figure}[t]
    \centering
    \includegraphics[width=0.99\linewidth]{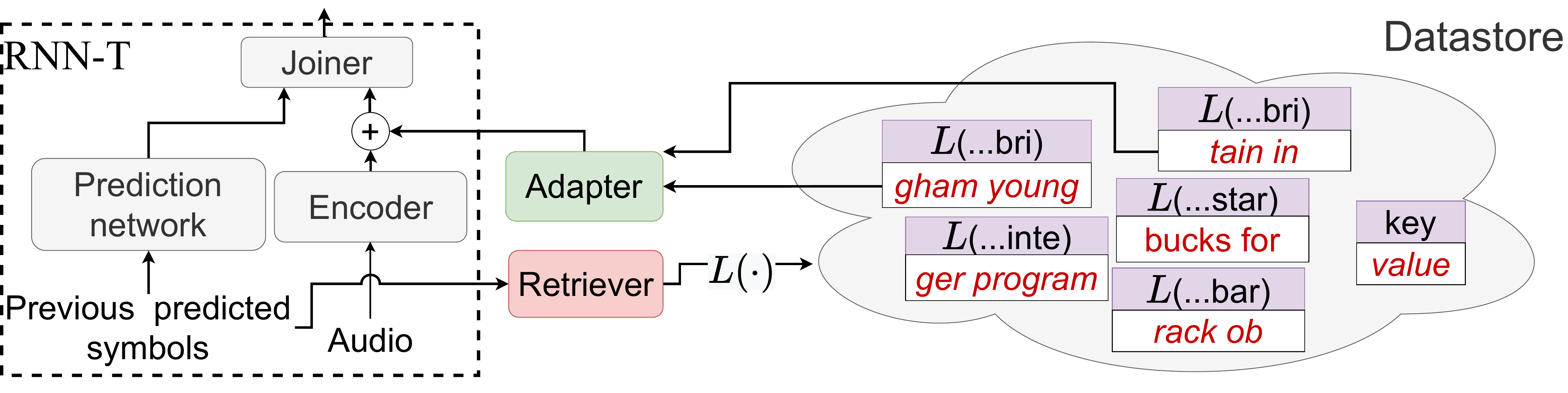}
    \caption{RNN-T modified to use retrieval from a datastore. ${L}(\dots \text{\textit{xyz}})$ denotes the retrieval LM encoding of some sentence ending in \textit{xyz}.}
    \label{fig:retrieve_and_adapt}
    \vspace{-2.5mm}
\end{figure}

\begin{figure*}[t]
    \centering
    \includegraphics[width=0.92\linewidth]{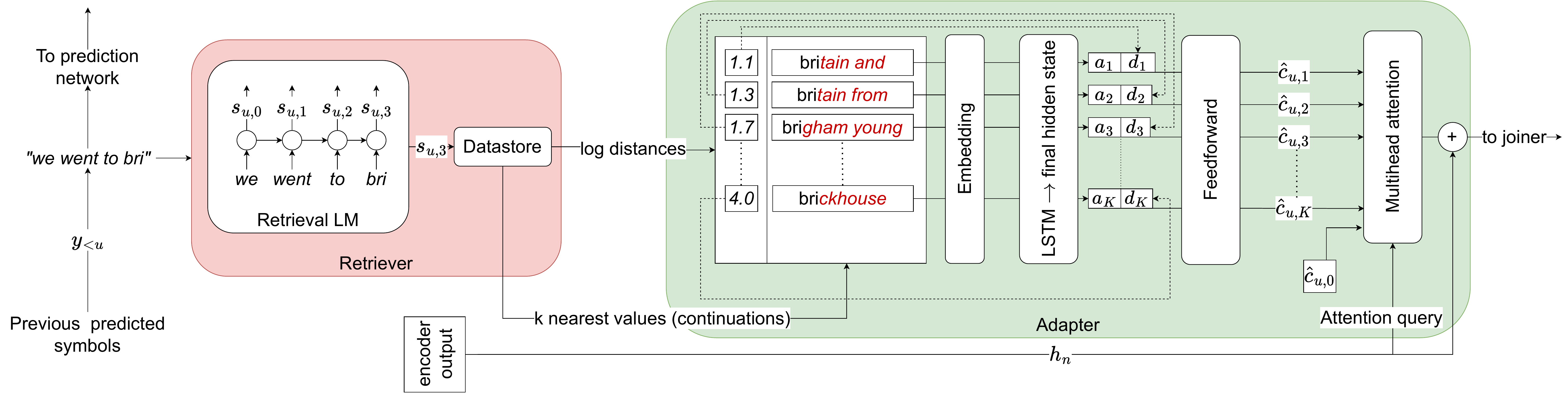}
    \caption{The retrieval LM computes an embedding for previously predicted symbols and retrieves the k nearest neighbors of this embedding are from a datastore. The continuations (italicized in red font) and log Euclidean distances of the neighbors are transformed by an encoding network and the resulting vectors are blended by an attention mechanism whose output is used to modify the original RNN-T encoder output.}
    \label{fig:retrieve_and_adapt_expand}
    \vspace{-2.5mm}
\end{figure*}
The model we propose is built on the RNN transducer~\cite{graves2012sequence}. The transducer is an end-to-end ASR model composed of three parts: the encoder, the prediction network and the joiner.

The encoder is a recurrent neural network which encodes a sequence of audio frames $(\Matrix{x}_1, \dots, \Matrix{x}_N)$ into hidden states $(\Matrix{h}_1, \dots, \Matrix{h}_N)$:
\useshortskip
\begin{align}
    \Matrix{h}_n = f^{enc}(\Matrix{x}_n, \Matrix{h}_{n-1}),
    \label{eqn:transcription}
\end{align}

The prediction network is a recurrent network which encodes the previously output non-blank tokens $y_{<u}$:
\useshortskip
\begin{align}
\Matrix{g}_u = f^{pred}(y_{u-1}, \Matrix{g}_{u-1}),
\label{eqn:prediction}
\end{align}

The joiner computes a joint embedding from the two outputs:
\useshortskip
\begin{align}
    \Matrix{z}_{n,u} = \phi(\Matrix{U} \Matrix{h}_n + \Matrix{V} \Matrix{g}_u + \Matrix{b_1}).
    \label{eqn:joiner}
\end{align}
$\Matrix{U}$, $\Matrix{V}$ and $\Matrix{b}_1$ are trainable parameters and $\phi$ is the hyperbolic tangent function.
The joint embedding is then used to compute a probability distribution over all tokens plus the blank token for alignment:
\useshortskip
\begin{align}
    p(y|n, u) = \sigma(\Matrix{W} \Matrix{z}_{n,u} + \Matrix{b}_2).
\end{align}
$\Matrix{W}$ and $\Matrix{b}_2$ are trainable parameters and $\sigma$ is the softmax function.

\subsection{Retrieval-augmented RNN-T}
Figure~\ref{fig:retrieve_and_adapt} depicts the modified RNN-T structure that we propose and
Figure~\ref{fig:retrieve_and_adapt_expand} illustrates the modifications in more detail.
Our modifications comprise a retriever which finds potential continuations for the RNN-T's current output from external text and an adapter which uses those continuations to bias the RNN-T's subsequent outputs.

\subsubsection{Retriever}
The retriever, depicted in pink in Figure~\ref{fig:retrieve_and_adapt_expand}, is based on the premise that embeddings generated by a pretrained neural language model for two similar phrases are closer in Euclidean space than those of two random phrases.
Therefore, to find potential continuations for any phrase (partial ASR hypotheses in our case) in a text corpus, we need to find phrases in that corpus that have similar embeddings to our phrase of interest, and return their continuations.

At the heart of our retriever is a pretrained LSTM LM which is used to generate embeddings for retrieval.\footnote{We use an LSTM instead of the transformers used in prior works on retrieval LM due to practical latency and memory considerations. Note that the retrieval LM needs not be trained on the text from which we retrieve.
In fact, in all our experiments, we use the same pretrained LSTM LM regardless of the adaptation text.
}
First we create a datastore for an adaptation text from which we intend to retrieve.
To do this, each sentence in the corpus is passed through the LM;
the LSTM's hidden state at each step is added as a key to the datastore,
with corresponding value comprising the input tokens to the next $t$ steps (we set $t=2$ in this paper).
By repeating this procedure for all sentences in the text, we get a key-value store, whose keys are the LM embeddings of phrases in the text, and whose values are the $t$-token long continuations of each key phrase.

To obtain candidate continuations of a partial ASR hypothesis during RNN-T decoding, we encode it with the retrieval LM and find the k nearest neighbors by Euclidean distance from the datastore.
Note that the query to the retrieval is the same as the input to the prediction network, i.e., the sequence of non-blank tokens ($y_{<u}$).
The retrieved continuations are then passed to the adapter along with the logarithms of the Euclidean distances between each key and the querying embedding.

\subsubsection{Adapter}
\label{subsection:adapter}
Having retrieved the k nearest neighbors, the question remains how best to integrate them back into the ASR.
Adopting the framework used for contextual biasing in~\cite{sathyendra2022contextual}, we introduce a trainable adapter to bias the RNN-T's encoder output before feeding it into the joint network.
The adapter, depicted in green in~Figure~\ref{fig:retrieve_and_adapt_expand}, comprises a recurrent encoder and a multihead dot-product attention mechanism.

The adapter encoder has an embedding layer which converts the tokens in each candidate continuation into dense form.
This is followed by an LSTM whose hidden state at the last step is taken as a fixed-length representation of that candidate.
The corresponding log-distance from the retriever is concatenated to give the model a clue about the relevance of each candidate, and this vector is further transformed by a feedforward network to get a final representation.

The multihead attention is used to select from the representations of the candidates.
The attention query is an affine projection of the encoder output $\Matrix{h}_n$\footnote{We also tried using $\Matrix{g}_u$ as the attention query. We found that while using $\Matrix{g}_u$ is computationally cheaper, using $\Matrix{h}_n$ results in better ASR performance.} and its keys and values are projections of the candidates' encodings.
Finally, the resulting context vector is added to the transcription output before passing the sum to the joiner.

In effect, the adapter modifies Equation~\ref{eqn:joiner} to:
\useshortskip
\begin{align}
    \Matrix{z}_{n,u} = \phi\Bigl(\Matrix{U} \bigl(\Matrix{h}_{n} + \sum_{k=0}^{K} \alpha_{n, u, k} \Matrix{c}_{u, k}\bigr) + \Matrix{V} \Matrix{g}_u + \Matrix{b}_1\Bigr),
    \label{eqn:adapt}
\end{align} 
where $K$ is the number of retrieved candidates and is a hyper-parameter,
$\Matrix{c}_{u, 1}, \dots, \Matrix{c}_{u, K}$ are the values of the attention mechanism computed from the retrieved candidates (by searching $y_{< u}$ in the datastore),
$\Matrix{c}_{u, 0} = \Matrix{c}$ is an extra trainable ``no-bias" embedding intended to give the model an option when all retrieved candidates are incorrect, and
$\alpha_{n, u, k}$ is the attention weight between $\Matrix{h}_n$ and $\Matrix{c}_{u, k}$.

%% file: inputfiles/reexperiments.tex
\subsection{Datasets}
\label{sec:experiments:datasets}
We experiment with two question-answering (QA) datasets: the Stanford Question Answering Dataset (Squad) v2.0~\cite{rajpurkar-etal-2018-know} and the CNN portion of the DeepMind Question Answering (DQA)~\cite{hermann2015teaching} dataset.
Each has a set of questions and a set of ``context" paragraphs which contain information useful for answering the questions.~\footnote{see {https://rajpurkar.github.io/SQuAD-explorer/explore/v2.0/dev} for examples of Squad contexts and associated questions.}

Our task is to perform ASR on the questions. We use the TTS system from~\cite{fazel21_interspeech} to obtain speech for the questions; and we construct datastores for retrieval from the contexts.
This setup emulates the data available to a typical voice assistant with open-source, non-proprietary data.
The synthesized questions correspond to user queries, and the contexts correspond to the knowledge base with which a downstream NLP module would resolve the queries.
Note that since we do not know exactly which context applies to which question, each dataset's datastore contains the keys and continuations of all contexts from the entire dataset.

\begin{table}[t]
    \centering
    \caption{Summary of the test sets. Squad-V and Squad-T refer to the Squad validation and test sets respectively, and DQA-V and DQA-T refer to the DQA validation and test sets respectively.}
    \begin{tabular}{lccccc}
        \toprule
        Dataset & \multicolumn{2}{c}{Contexts} & \multicolumn{2}{c}{Questions} \\
         & \#Paragraphs & \#Tokens & \#Sentences & \#Words \\
        \midrule
        Squad-V & 19124 & 3997k & 6510 & 68107 \\
        Squad-T & 1204 & 287k & 11868 & 123020 \\
        DQA-V & 1220 & 1371k & 3924 & 53494 \\
        DQA-T & 1093 & 1172k & 3198 & 44737 \\
        \bottomrule
    \end{tabular}
    \label{tab:data_stats}
\end{table}

In Table~\ref{tab:data_stats}, we report the number of context paragraphs, questions and their constituent tokens and words for each dataset.
Note that for Squad, since the official test set is not public, we use the official dev-set as our test set (Squad-T), and split off $5\%$ of the training questions for validation (Squad-V).
We therefore use the entire set of Squad training contexts for the Squad-V datastore, which makes the datastore large and retrieval that much harder.
For DQA, all entities are de-anonymized before TTS and datastore construction.

To get a strong DQA baseline, we pretrain the baseline RNN-T for 100k steps on a mix of the DQA and Librispeech~\cite{panayatov2015librispeech} training sets, with batches uniformly sampled from each set. The DQA training set contains 380k questions which--to avoid any acoustic mismatch--we synthesize with the same TTS system as the test sets, using speaker profiles which include those used for the test sets.
Compared to a baseline trained purely on Librispeech, including the DQA training set improves the word error rates (WER) from 34.3\% to 13.8\% on DQA-T, from 23.5\% to 15.2\% on Squad-T, with only a negligible degradation on the Librispeech test sets (6.5\% to 6.6\% on the test-clean).

With the parameters of the pretrained RNN-T frozen, we train the adapter for 30k steps on the Squad training set, which contains 124k sentences (the official training set minus the validation sentences); we use its synthesized questions as ASR data along with a datastore of all its contexts (same datastore as Squad-V).

To mitigate the risk of adapter overfitting on retrieved continuations, we add two forms of noise and force the model to learn when to rely on the no-bias embedding.
First, we mix in an equal proportion of Librispeech batches to the \emph{adapter} training data so that the retrieved continuations from the Squad datastore would be irrelevant.
In addition, with $0.1$ probability we retrieve random continuations from the training datastore instead of the actual k nearest neighbors, so that the model is exposed to--and learns to deal with--incorrect continuations even when the domain matches.

\subsection{Model configuration}
The baseline RNN-T has 64 million parameters, comprising an encoder with six LSTM layers with 1024 units followed by a 640-dimensional affine projection,
a prediction network which has two LSTM layers of 1024 units also followed by projection to 640 dimensions,
and a joiner with intermediate dimension of 512 and output dimension of 2501 corresponding to 2500 unigram subword tokens~\cite{kudo2018subword} trained on Librispeech, and the extra blank token.

The adapter adds 1 million parameters, of which $2501 \times 128 \approx 320k$ are in the embedding layer.
The remaining parameters are in two 128-unit LSTM layers, two feedforward layers with 128 units and multihead attention with two heads and key dimension of 128.

The retrieval LM is a two-layer LSTM with 256 units trained on Wikitext-103~\cite{merity2016pointer}.
We reiterate that this retrieval LM is kept fixed regardless of which datastore we retrieve from.

We implement the k-nearest-neighbor search in FAISS~\cite{johnson2019billion} using a CPU index with product quantization~\cite{jegou2010product}.
The largest index in our experiments--constructed by concatenating all datastores (used in the ``All" rows of Tables~\ref{tab:main_results}~and~\ref{tab:ne_results})--occupies just under 500 megabytes of RAM.
We set $K$ to 16 for both training and inference, i.e. at each step, we retrieve 16 candidate continuations out of hundreds of thousands to millions (number context tokens in Table~\ref{tab:data_stats}).

\subsection{Test set performance}
\label{sec:experiments:test}
Table~\ref{tab:main_results} shows the word error rates of the various test sets.

Since our retrieval mechanism involves introducing and training an adapter, some of the improvement or degradation in performance may be attributed to simply having extra parameters trained on question-answering data, essentially updating the RNN-T's internal language model (ILM), rather than being able to retrieve and use the correct continuations.
To measure this effect, we train another baseline which has an adapter but no retriever.
For this, we input to the adapter LSTM a single trainable embedding instead of the embeddings of retrieved continuations.
This ``fixed embedding" approach, when compared to the baseline without any adaptations, improves Squad and degrades DQA performance.
This is expected because the baseline training includes DQA training data, while the adapter is trained with Squad and Librispeech ASR data.

We observe significant improvements compared to either baseline on all test sets when the datastore \emph{matches} the corresponding contexts, e.g. Squad-T datastore for Squad-T test set etc.
It is noteworthy that we get relative improvements of $9\%$ and $7\%$ respectively on the DQA validation and test sets compared to the baseline with no adapter despite the performance drop that we incur due to the ILM shift from training the adapter on Squad+Librispeech (as evidenced by the fixed embedding results).

Next, we observe that the performance improvements brought by retrieval are generally proportional to how relevant the datastore is. For instance, when decoding the Squad test set, the WER increases when we switch from the datastore of Squad test contexts to the datastore of Squad validation contexts and increases further as we switch to the datastores of DQA contexts, at which point we get performance comparable to using the fixed embedding.

The matched results are predicated on picking the right datastore for each test utterance.
We also consider retrieving from a single large datastore containing contexts from all the datasets (the ``All" rows in the table).
While this performs worse than picking the matching datastore, it is significantly better than using any other single datastore.
This implies that even in the presence of a few million extra distracting contexts in the datastore, the retriever still does a good job of retrieving the correct ones.
Thus, in practice, it would be a viable strategy to concatenate several data stores unless sure of which one to pick.

The second partition of the table shows the results of using shallow fusion on each system.
For the DQA test sets, we use an LSTM LM trained on the DQA training contexts for shallow fusion.
For Squad, we use an LSTM LM trained on Wikitext-103 (the same one used as the retrieval LM), since both Squad and Wikitext are constructed from Wikipedia data, and the Squad dataset is too small to train a robust LM.
This reflects one of the advantages of retrieval: we can add any relevant corpus, no matter how small, to the datastore without danger of overfitting to it.
We optimize the LM interpolation weights separately on each validation set and apply them to the corresponding test sets.
Retrieval is comparable to shallow fusion on DQA test sets and outperforms it on Squad. 
Furthermore, the two are complementary as further significant improvements can be obtained by combining retrieval with shallow fusion.

The final partition shows the result of retrieving not from a datastore of contexts but of the questions themselves (with an adapter trained for questions).
This gives us an upper-bound, however unrealistic, on performance in the ``Match" setting.
We observe that by retrieving from datastores of questions, we can more than halve the WER to around 5\% across all test sets.
This indicates that even though we only retrieve 16 continuations at a time out of hundreds of thousands (after tokenization of \#Words in Table~\ref{tab:data_stats}), the main bottleneck is not in the retrieval itself, but the simple fact that the contexts and the questions are not perfectly matched.
The residual WER tells us the inherent errors due to either the retriever failing to retrieve the correct continuations or the adapter failing to bias the ASR output.

\begin{table}[t]
    \centering
    \caption{WER on various test sets as the datastore is varied compared to the baseline with no retrieval (``None") and a baseline with retrieval replaced by a fixed embedding (``Fixed emb."). S-F denotes the use of shallow fusion.}
    \begin{tabular}{lccccc}
        \toprule
        Datastore & S-F & Squad-V & Squad-T & DQA-V & DQA-T  \\
        \midrule
        None & \ding{55} & 15.8 & 15.2 & 14.0 & 13.8 \\
        Fixed emb. & \ding{55} & 14.4 & 13.8 & 14.7 & 14.7 \\
        Squad-V & \ding{55} & \textbf{11.9} & 12.5 & 15.1 & 15.1 \\
        Squad-T & \ding{55} & 13.6 & \textbf{11.1} & 15.5 & 15.7 \\
        DQA-V & \ding{55} & 14.2 & 13.6 & \textbf{12.7} & 14.3 \\
        DQA-T & \ding{55} & 14.4 & 13.8 & 14.6 & \textbf{12.8} \\
        All &  \ding{55} & 12.3 & 11.9 & 13.3 & 13.3 \\
        \midrule
        None & \ding{51} & 13.7 & 13.1 & 12.5 & 12.4 \\
        Fixed emb. & \ding{51} & 12.3 & 11.7 & 13.4 & 13.3 \\
        Match & \ding{51} & \textbf{11.0} & \textbf{9.9} & \textbf{11.7} & \textbf{11.8} \\
        All & \ding{51} & 11.4 & 10.8 & 12.4 & 12.5 \\
        \midrule
        Questions & \ding{55} & 4.5 & 4.5 & 5.5 & 5.8 \\
        \bottomrule
    \end{tabular}
    \label{tab:main_results}
\end{table}

\subsection{Performance on named entities}
Table~\ref{tab:ne_results} shows the results on the DQA test sets split by whether or not the reference word is a named entity.
We report results only on DQA because, unlike DQA, the Squad dataset references have no named entity tags.
We observe that retrieval generally improves named entities more than it does other words.
Adding retrieval to the baseline RNN-T leads to relative WER improvements of $11\%$ and $8\%$ respectively on named entities and other words on the validation set. The respective improvements on the test set are $12\%$ and $4\%$.
We observe that shallow fusion improves more on regular words and less on named entities compared to retrieval.
Finally, when we use shallow fusion and retrieval together, we get better named entity WER that using either by itself, but the WER on other words does not get better than using shallow fusion by itself.
This supports our thesis that the trained shallow fusion LM can do fine by itself for common words, and the utility of retrieval is most pronounced for rare words.

\begin{table}[t]
    \centering
    \caption{DQA dataset WERs for named entities and common words.}
    \begin{tabular}{lccccc}
        \toprule
        Datastore & S-F & \multicolumn{2}{c}{DQA-V} & \multicolumn{2}{c}{DQA-T}  \\
        & & Entities & Others & Entities & Others \\
        \midrule
        None & \ding{55} & 25.8 & 10.0 & 27.1 & 9.8 \\
        Fixed emb. & \ding{55} & 27.1 & 10.6 & 28.5 & 10.4 \\
        Match & \ding{55} & \textbf{23.0} & \textbf{9.2} & \textbf{23.9} & \textbf{9.4} \\
        All & \ding{55} & 24.1 & 9.7 & 24.9 & 9.8 \\
        \midrule
        None & \ding{51} & 24.6& \textbf{8.5} & 25.6  & \textbf{8.4} \\
        Fixed emb. & \ding{51} & 25.2 & 9.5 & 26.7 & 9.2 \\
        Match & \ding{51} & \textbf{21.4} & \textbf{8.5} & \textbf{22.4} & {8.6} \\
        All & \ding{51} & 22.7 & 9.0 & 24.6 & 9.0 \\
        \bottomrule
    \end{tabular}
    \label{tab:ne_results}
\end{table}

\subsection{Impact of number of retrieved neighbors}
\begin{table}[t]
    \setlength\tabcolsep{4.5pt}
    \centering
    \caption{WER obtained by retrieving from the mixed datastore as the number of retrieved neighbors is varied.}
    \begin{tabular}{lcccccccc}
    \toprule
         Test set & - & 1 & 2 & 4 & 8 & 16 & 32 & 64  \\
         \midrule
         Squad-V & 15.8 & 16.2 & 15.1 & 13.4 & 12.7 & 12.3 & 12.2 & \textbf{12.0} \\
         Squad-T & 15.2 & 16.3 & 14.8 & 13.3 & 12.4 & 11.9 & 11.6 & \textbf{11.5} \\
         DQA-V & 14.0 & 17.0 & 15.0 & 14.1 & 13.7 & 13.3 & 13.1 & \textbf{12.9} \\
         DQA-T & 13.8 & 16.4 & 14.9 & 13.8 & 13.5 & 13.3 & 13.2 & \textbf{13.0} \\
         \bottomrule
    \end{tabular}
    \label{tab:knn_mix}
\end{table}
Table~\ref{tab:knn_mix} shows the WERs on the test sets as we vary $K$ while retrieving from the datastore of all datasets (``All").
Note that we only vary $K$ at inference time; during training it is still fixed to 16.
We observe that the WER improves--with diminishing returns--as $K$ increases.

In experiments whose results we omit due to space constraints, we varied $K$ at training time and note that the higher we set $K$ at training time, the higher we can, and have to, set it during inference. To get improved the results with lower values of $K$ at inference, we need to train with low values of $K$.
For instance, across test sets, the WERs obtained from setting $K=1$ for both training and testing fall between those obtained from setting $K=4$ and $K=8$ in Table~\ref{tab:knn_mix}.

%% file: inputfiles/conclusion.tex
In this work, we have proposed augmenting an RNN-T based ASR model with a retrieval mechanism, which searches an external datastore for potential continuations of partial ASR hypotheses.
We show that biasing subsequent decoding steps with these continuations significantly improves ASR performance, especially on named entities, when the datastore contains relevant text.
We further show that retrieval can be complemented by a conventional shallow fusion LM,
as using the two in tandem results in further improvements.

Avenues for future work include replacing the retrieval LM with a model trained explicitly for retrieval,
further reducing performance degradation when the datastore and test set are unrelated,
and improving efficiency by exploring ways to reduce retrieval frequency.

%% file: inputfiles/extras.tex
This appendix contains experimental results that were not included in the main paper due to ICASSP space constraints.

\subsection{Impact of including DQA training data in baseline RNN-T training}
Table~\ref{tab:to_finetune_or_not_and_where} shows the impact of including DQA data when training the RNN-T baseline.
The first group of results uses Librispeech and DQA data for training the baseline RNN-T (these are the results reported in the main body of the paper), while the second group has a baseline trained on Librispeech data only.
As before, the retriever is always trained on the Squad training set.

Adding DQA training data improves the baseline considerably, since it better matches the acoustics of the test sets.
Therefore, we took this baseline in the paper.
Nevertheless, for the baseline trained with only Librispeech data, retrieval leads to significant reductions in word error rates--even more so than for the other baseline.
Note that this is despite the higher retrieval quality discrepancy between training and inference since higher word error rates lead to more noise in the retrieved tokens.

All subsequence subsections report results on the validation sets with the baseline trained on Librispeech only.
\begin{table}[b]
    \centering
    \setlength\tabcolsep{4.5pt}
    \caption{WER with retrieval on baselines trained with or without DQA training data. L + D denotes the baseline trained with a mix of Librispeech and DQA, while L denotes the baseline trained with only Librispeech data. $^*$ denotes the setting used in the main paper.}
    \begin{tabular}{lccccc}
        \toprule
        Datastore & Baseline & Squad-V & Squad-T & DQA-V & DQA-T  \\
        \midrule
        None & L + D$^*$ & 15.8 & 15.2 & 14.0 & 13.8 \\
        Fixed emb. & L + D$^*$ & 14.4 & 13.8 & 14.7 & 14.7 \\
        Match & L + D$^*$ & 11.9 & 11.1 & 12.7 & 12.8 \\
        All & L + D$^*$ & 12.3 & 11.9 & 13.3 & 13.3 \\
        \midrule
        None & L & 24.7 & 23.5 & 35.2 & 34.3 \\
        Fixed emb. & L & 20.4 & 18.9 & 33.1 & 33.1 \\
        Match & L & 15.5 & 14.2 & 27.4 & 26.4 \\
        All & L & 16.2 & 14.8 & 27.1 & 27.1 \\
        \bottomrule
    \end{tabular}
    \label{tab:to_finetune_or_not_and_where}
\end{table}

\subsection{Adapting encoder output vs prediction network output}
In our experiments, we used the retrieved tokens to adapt the encoder output.
We also experiment with adapting the prediction network output instead.
Essentially, instead of~\eqref{eqn:adapt}, we use:
\begin{align}
    \Matrix{z}_{n,u} = \phi\Bigl(\Matrix{U}\Matrix{h}_{n} + \Matrix{V}\bigl( \Matrix{g}_u + \sum_{k=0}^{K} \alpha_{u, k} \Matrix{c}_{u, k} \bigr) + \Matrix{b}_1\Bigr),
    \label{eqn:adapt_prediction}
\end{align} 
where the attention weights $\{\alpha_{u, k}\}$ are computed with the prediction network output $\Matrix{g}_u$ as the attention query.
This leads to a computationally cheaper system than the one querying with the encoder output--note the absence of a temporal index ($n$) in the attention weights here compared to $\{\alpha_{n, u, k}\}$ in ~\eqref{eqn:adapt}.

Table~\ref{tab:what_to_adapt} shows the result of adapting the various components on the validation sets while also varying the datastore from which continuations are retrieved.
When the adaption involves a single trained embedding, both choices result in almost identical performance.
With actual retrieval, while both adaption choices significantly improve the baseline, adapting the encoder results in much lower WER compared to adapting the prediction network output.
There is therefore a trade-off between accuracy and computational cost as adapting the encoder is better in terms of WER but much costlier due to the extra attention dimension.
\begin{table}[t]
    \centering
    \setlength\tabcolsep{4.5pt}
    \caption{WER impact of adapting the encoder or prediction network outputs. Results with ``f.e." denote the baseline which use fixed embeddings instead of actual retrieval. $^*$ denotes the setting used in the rest of the paper.}
    \begin{tabular}{lccccc}
    \toprule
         Test set & \multicolumn{2}{c}{Squad-V} && \multicolumn{2}{c}{DQA-V} \\
         Datastore & Squad-V & DQA-V && Squad-V & DQA-V \\
         \midrule
         Adaptee \\
         \midrule
         - & 24.7 & 24.7 && 35.2 & 35.2 \\
         \midrule
         Prediction f.e. & 20.4 & 20.4 && 33.6 & 33.6 \\
         Encoder f.e. & 20.4 & 20.4 && 33.1 & 33.1 \\
         \midrule
         Prediction & 16.7 & 18.7 && 31.3 & 28.6 \\
         Encoder$^*$ & 15.5 & 18.1 && 29.5 & 27.4 \\
         \bottomrule
    \end{tabular}
    \label{tab:what_to_adapt}
\end{table}

\subsection{Impact of retrieval noise in training}
\begin{table}[b]
    \centering
    \setlength\tabcolsep{4.5pt}
    \caption{WER as retrieval noise during adapter training is varied. $^*$~denotes the setting used in the rest of the paper.}
    \begin{tabular}{lcccccc}
    \toprule
         Test set && \multicolumn{2}{c}{Squad-V} && \multicolumn{2}{c}{DQA-V} \\
         Datastore && Squad-V & DQA-V && Squad-V & DQA-V \\
         \midrule
         Prob. & +Libri \\
         \midrule
         0 &\ding{55}& 15.6 & 20.3 && 31.0 & 28.0 \\
         0.1 &\ding{55}& 15.0 & 18.3 && 30.0 & 26.6 \\
         0.3 &\ding{55}& 14.6 & 16.9 && 28.9 & 26.7 \\
         0.5 &\ding{55}& 15.0 & 16.5 && 28.8 & 26.9 \\
         \midrule
         0 &\ding{51}& 15.5 & 18.9 && 29.9 & 27.0 \\
         0.1$^*$ &\ding{51}& 15.5 & 18.1 && 29.5 & 27.4 \\
         0.3 &\ding{51}& 15.5 & 17.3 && 29.3 & 26.8 \\
         0.5 &\ding{51}& 15.9 & 17.1 && 29.1 & 28.3 \\
         \bottomrule
    \end{tabular}
    \label{tab:retrieval_noise}
\end{table}
As described in the last paragraph of Section~\ref{sec:experiments:datasets}, we add some noise to the adapter training to avoid model overconfidence; we do this by sampling Librispeech training data (for which the training datastore has no relevant entries) with 0.5 probability and retrieving with a further 0.1 probability random entries from the datastore instead of the actual k nearest neighbors. Table~\ref{tab:retrieval_noise} shows the WER impact of these.

First, we note that regardless of whether Librispeech utterances are included in the adapter training, increasing the probability of random retrievals always improves the WER in the cross-retrieval setting, i.e. retrieving from Sqaud-V datastore when decoding DQA-V utterances and vice versa.
Thus, training with random retrieval serves its intended purpose: mitigating the performance degradation when the retrieval datastore is not related to the test set.
The cost of course is that as the noise probability is increased beyond some threshold ($0.3$ in these experiments), the WER in the matched setting--using Squad datastore for Squad test set and DQA for DQA--starts to deteriorate.

Finally, we note that adding noise in the form of Librispeech data also has a similar effect, especially when the probability of random retrieval is at or below $0.1$.
As the probability of random retrieval increases, using Librispeech data starts to hurts performance compared to not using it.

\subsection{Using pretrained encoder in the adapter}
The adapter comprises a recurrent encoder and a multihead attention which are jointly trained.
Here we experiment with using a frozen pretrained encoder, and only train the multihead attention and feedforward part of the encoder.
Specifically, we use the retrieval LSTM LM to encode the continuations.
This significantly improves training throughput (utterances/second) as it allows offline computation of the training embeddings.

Table~\ref{tab:trained_or_frozen} shows a WER comparison of the two approaches.
While using the pretrained encoder works better than the baselines without retrieval, it is significantly worse than retrieval with trained encoder.
Although finetuning the pretrained encoder is another option, it would defeat our purpose of increasing training throughput.
It may still be worth exploring if such a finetuning approach leads to better results or faster convergence.

\begin{table}[t]
    \centering
    \caption{WER comparsion between training the adapter's recurrent encoder or using the retrieval LM for encoding the continuations.}
    \setlength\tabcolsep{4.5pt}
    \begin{tabular}{lccccc}
    \toprule
         Test set & \multicolumn{2}{c}{Squad-V} && \multicolumn{2}{c}{DQA-V} \\
         Datastore & Squad-V & DQA-V && Squad-V & DQA-V \\
         \midrule
         Embedding \\
         \midrule
          - & 24.7 & 24.7 && 35.2 & 35.2 \\
          Fixed embedding & 20.4 & 20.4 && 33.1 & 33.1 \\
          \midrule
          Trained$^*$& 15.5 & 18.1 && 29.5 & 27.4 \\
          Pretrained & 16.9 & 19.6 && 30.6 & 31.1 \\
          \bottomrule
    \end{tabular}
    \label{tab:trained_or_frozen}
\end{table}

\subsection{Number of attention heads}
In our experiments, we have set the number of attention heads in the adapter to 2.
Table~\ref{tab:num_heads} compares this to setting the it to 1.

Setting the number of heads to 1 slightly degrades the word error rate.
However, this may be worth it due the computational savings, especially at inference time where it essentially halves the computational cost of the adapter--since the cost of embedding the continuations can be amortized away by offline pre-computation.
\begin{table}[t]
    \centering
    \def\arraystretch{0.88}
    \setlength\tabcolsep{4.5pt}
    \caption{WER as the number of attention heads is reduced}
    \begin{tabular}{lccccc}
    \toprule
         Test set & \multicolumn{2}{c}{Squad} && \multicolumn{2}{c}{DQA} \\
         Datastore & Squad & DQA && Squad & DQA \\
         \midrule
         Num heads \\
         \midrule
         1 & 15.9 & 18.5 && 30.2 & 27.4 \\
         2$^*$ & 15.5 & 18.1 && 29.5 & 27.4 \\
         \bottomrule
    \end{tabular}
    \label{tab:num_heads}
\end{table}

\subsection{Impact of varying the number of retrieved tokens}
\begin{table}[t]
    \centering
    \setlength\tabcolsep{4.5pt}
    \caption{WER on the Squad-V set as K is varied while retrieving from the Squad-V datastore.}
    \npdecimalsign{.}
    \nprounddigits{2}
    \begin{tabular}{lccccccc}
    \toprule
         Test & 1 & 2 & 4 & 8 & 16 & 32 & 64  \\
         Train \\
         \midrule
         1 & 16.6 & 16.5 & 16.5 & 17.1 & 18.1 & 19.5 & 20.9 \\
         2 & 17.9 & 16.6 & 16.1 & 16.2 & 16.4 & 16.8 & 17.3 \\
         4 & 19.2 & 17.5 & 16.1 & 15.8 & 15.8 & 15.9 & 16.0 \\
         8 & 19.9 & 18.6 & 16.8 & 15.9 & 15.7 & 15.6 & 15.4 \\
         16 & 24.2 & 19.5 & 17.4 & 16.3 & 15.5 & 15.5 & 15.4 \\
         \bottomrule
    \end{tabular}
    \npnoround
    \label{tab:knn_squad_squad}
\end{table}

\begin{table}[t]
    \centering
    \setlength\tabcolsep{4.5pt}
    \caption{WER on the Squad-V set as K is varied while retrieving from the DQA-V datastore.}
    \begin{tabular}{lccccccc}
    \toprule
         Test & 1 & 2 & 4 & 8 & 16 & 32 & 64  \\
         Train \\
         \midrule
         1 & 17.9 & 17.6 & 17.5 & 18.0 & 18.3 & 18.7 & 19.0 \\
         2 & 18.7 & 17.9 & 17.6 & 17.5 & 17.5 & 17.7 & 17.9 \\
         4 & 19.6 & 19.0 & 18.3 & 17.9 & 17.5 & 17.5 & 17.6 \\
         8 & 21.9 & 20.7 & 19.7 & 17.9 & 17.5 & 17.1 & 17.0 \\
         16 & 22.3 & 21.4 & 19.5 & 18.6 & 18.1 & 17.6 & 17.2 \\
         \bottomrule
    \end{tabular}
    \label{tab:knn_squad_dqa}
\end{table}

\begin{table}[t!]
    \centering
    \setlength\tabcolsep{4.5pt}
    \caption{WER on the DQA-V set as K is varied while retrieving from the Squad-V datastore.}
    \begin{tabular}{lccccccc}
    \toprule
         Test & 1 & 2 & 4 & 8 & 16 & 32 & 64  \\
         Train \\
         \midrule
         1 & 30.5 & 30.5 & 30.3 & 30.5 & 31.6 & 32.6 & 33.8 \\
         2 & 32.1 & 30.6 & 30.5 & 30.6 & 30.7 & 30.8 & 31.5 \\
         4 & 32.5 & 31.0 & 30.1 & 29.8 & 29.4 & 29.5 & 29.5 \\
         8 & 33.1 & 31.9 & 30.3 & 29.7 & 29.5 & 28.9 & 28.8 \\
         16 & 35.7 & 33.2 & 31.4 & 30.5 & 29.5 & 29.0 & 29.0 \\
         \bottomrule
    \end{tabular}
    \label{tab:knn_dqa_squad}
\end{table}

\begin{table}[t!]
    \centering
    \setlength\tabcolsep{4.5pt}
    \caption{WER on the DQA-V set as K is varied while retrieving from the DQA-V datastore.}
    \begin{tabular}{lccccccc}
    \toprule
         Test & 1 & 2 & 4 & 8 & 16 & 32 & 64  \\
         Train \\
         \midrule
         1 & 28.6 & 28.8 & 28.5 & 29.1 & 29.0 & 29.7 & 30.4 \\
         2 & 29.9 & 29.3 & 28.5 & 28.4 & 28.2 & 28.3 & 28.5\\
         4 & 29.5 & 28.6 & 28.1 & 27.6 & 27.2 & 27.0 & 26.9 \\
         8 & 30.1 & 28.9 & 28.0 & 27.1 & 26.8 & 26.7 & 26.5 \\
         16 & 31.1 & 30.3 & 28.5 & 27.9 & 27.4 & 27.2 & 27.2 \\
         \bottomrule
    \end{tabular}
    \label{tab:knn_dqa_dqa}
\end{table}
Tables~\ref{tab:knn_squad_squad}~to~\ref{tab:knn_dqa_dqa} show the word error rates for different datastore/test set combinations as the number of retrieved continuations is varied at training time and during inference.

The best word error rates tend towards the bottom right corner of each table--where $K$ is high for both training and inference. The worst are on the bottom left followed by the top-right--where there is severe mismatch between the number of $K$ at training time and at test time. The top-left corners--where $K$ is low for both training and inference--have word error rates in between.

Generally, the WER improves as $K$ is increased at inference time up to around 4 times the setting during training after which it starts to degrade.
Conversely, performance degrades rapidly--in some cases, approaching the baseline performance--when $K$ is set low during inference relative to training.

This indicates that $K$ should be set at training time to match the intended setting for inference to avoid degradation in performance.
An alternative training strategy would be not to fix $K$ during training but rather to sample a different $K$ for, say, each training batch.

%% file: Template.bbl
\begin{thebibliography}{10}

\bibitem{wang21t_interspeech}
Peidong Wang, Tara~N. Sainath, and Ron~J. Weiss,
\newblock ``{Multitask Training with Text Data for End-to-End Speech
  Recognition},''
\newblock in {\em Proc. Interspeech 2021}, 2021, pp. 2566--2570.

\bibitem{yusuf22usted}
Bolaji Yusuf, Ankur Gandhe, and Alex Sokolov,
\newblock ``{USTED: Improving ASR with a Unified Speech and Text
  Encoder-Decoder},''
\newblock in {\em Proceedings of ICASSP 2022 - 2022 IEEE International
  Conference on Acoustics, Speech and Signal Processing (ICASSP)}. 2022, pp.
  8297--8301, IEEE Signal Processing Society.

\bibitem{thomas2022integrating}
Samuel Thomas, Brian Kingsbury, George Saon, and Hong-Kwang~J Kuo,
\newblock ``{Integrating Text Inputs For Training and Adapting RNN Transducer
  ASR Models},''
\newblock in {\em ICASSP 2022-2022 IEEE International Conference on Acoustics,
  Speech and Signal Processing (ICASSP)}. IEEE, 2022, pp. 8127--8131.

\bibitem{gulcehre2015using}
Caglar Gulcehre et~al.,
\newblock ``On using monolingual corpora in neural machine translation,''
\newblock {\em arXiv preprint arXiv:1503.03535}, 2015.

\bibitem{Chorowski2017}
Jan Chorowski and Navdeep Jaitly,
\newblock ``Towards better decoding and language model integration in sequence
  to sequence models,''
\newblock in {\em Proc. Interspeech 2017}, 2017, pp. 523--527.

\bibitem{variani2020hybrid}
Ehsan Variani, David Rybach, Cyril Allauzen, and Michael Riley,
\newblock ``{Hybrid Autoregressive Transducer (HAT)},''
\newblock in {\em ICASSP 2020-2020 IEEE International Conference on Acoustics,
  Speech and Signal Processing (ICASSP)}. IEEE, 2020, pp. 6139--6143.

\bibitem{mcdermott2019density}
Erik McDermott, Hasim Sak, and Ehsan Variani,
\newblock ``{A Density Ratio Approach to Language Model Fusion in End-to-End
  Automatic Speech Recognition},''
\newblock in {\em 2019 IEEE Automatic Speech Recognition and Understanding
  Workshop (ASRU)}, 2019, pp. 434--441.

\bibitem{meng2021internal}
Zhong Meng et~al.,
\newblock ``Internal language model training for domain-adaptive end-to-end
  speech recognition,''
\newblock in {\em ICASSP 2021-2021 IEEE International Conference on Acoustics,
  Speech and Signal Processing (ICASSP)}. IEEE, 2021, pp. 7338--7342.

\bibitem{zhao2019shallow}
Ding Zhao et~al.,
\newblock ``{Shallow-Fusion End-to-End Contextual Biasing},''
\newblock in {\em Interspeech}, 2019, pp. 1418--1422.

\bibitem{gourav2021personalization}
Aditya Gourav et~al.,
\newblock ``Personalization strategies for end-to-end speech recognition
  systems,''
\newblock in {\em ICASSP 2021-2021 IEEE International Conference on Acoustics,
  Speech and Signal Processing (ICASSP)}. IEEE, 2021, pp. 7348--7352.

\bibitem{kocour2021interspeech}
Martin Kocour et~al.,
\newblock ``{Boosting of Contextual Information in ASR for Air-Traffic
  Call-Sign Recognition},''
\newblock in {\em Proceedings Interspeech 2021}. 2021, vol. 2021, pp.
  3301--3305, International Speech Communication Association.

\bibitem{pundak2018deep}
Golan Pundak, Tara~N Sainath, Rohit Prabhavalkar, Anjuli Kannan, and Ding Zhao,
\newblock ``Deep context: end-to-end contextual speech recognition,''
\newblock in {\em 2018 IEEE spoken language technology workshop (SLT)}. IEEE,
  2018, pp. 418--425.

\bibitem{jain20_interspeech}
Mahaveer Jain, Gil Keren, Jay Mahadeokar, Geoffrey Zweig, Florian Metze, and
  Yatharth Saraf,
\newblock ``{Contextual RNN-T for Open Domain ASR},''
\newblock in {\em Proc. Interspeech 2020}, 2020, pp. 11--15.

\bibitem{feng2021}
Feng-Ju Chang et~al.,
\newblock ``{Context-Aware Transformer Transducer for Speech Recognition},''
\newblock in {\em 2021 IEEE Automatic Speech Recognition and Understanding
  Workshop (ASRU)}, 2021, pp. 503--510.

\bibitem{sathyendra2022contextual}
Kanthashree~Mysore Sathyendra et~al.,
\newblock ``Contextual adapters for personalized speech recognition in neural
  transducers,''
\newblock in {\em ICASSP 2022-2022 IEEE International Conference on Acoustics,
  Speech and Signal Processing (ICASSP)}. IEEE, 2022, pp. 8537--8541.

\bibitem{dingliwal2022}
Saket Dingliwal, Monica Sunkara, Srikanth Ronanki, Jeff Farris, Katrin
  Kirchhoff, and Sravan Bodapati,
\newblock ``Personalization of ctc speech recognition models,''
\newblock in {\em IEEE 2022 Workshop on Spoken Language Technology}, 2022.

\bibitem{khandelwal2020generalization}
Urvashi Khandelwal, Omer Levy, Dan Jurafsky, Luke Zettlemoyer, and Mike Lewis,
\newblock ``{Generalization through Memorization: Nearest Neighbor Language
  Models},''
\newblock in {\em International Conference on Learning Representations (ICLR)},
  2020.

\bibitem{guu2020retrieval}
Kelvin Guu, Kenton Lee, Zora Tung, Panupong Pasupat, and Mingwei Chang,
\newblock ``Retrieval augmented language model pre-training,''
\newblock in {\em International Conference on Machine Learning}. PMLR, 2020,
  pp. 3929--3938.

\bibitem{he2021efficient}
Junxian He, Graham Neubig, and Taylor Berg-Kirkpatrick,
\newblock ``Efficient nearest neighbor language models,''
\newblock in {\em Proceedings of the 2021 Conference on Empirical Methods in
  Natural Language Processing}, 2021, pp. 5703--5714.

\bibitem{rajpurkar-etal-2018-know}
Pranav Rajpurkar, Robin Jia, and Percy Liang,
\newblock ``Know what you don{'}t know: Unanswerable questions for {SQ}u{AD},''
\newblock in {\em Proceedings of the 56th Annual Meeting of the Association for
  Computational Linguistics (Volume 2: Short Papers)}, Melbourne, Australia,
  July 2018, pp. 784--789, Association for Computational Linguistics.

\bibitem{hermann2015teaching}
Karl~Moritz Hermann, Tomas Kocisky, Edward Grefenstette, Lasse Espeholt, Will
  Kay, Mustafa Suleyman, and Phil Blunsom,
\newblock ``Teaching machines to read and comprehend,''
\newblock {\em Advances in neural information processing systems}, vol. 28,
  2015.

\bibitem{graves2012sequence}
Alex Graves,
\newblock ``Sequence transduction with recurrent neural networks,''
\newblock {\em arXiv preprint arXiv:1211.3711}, 2012.

\bibitem{fazel21_interspeech}
Amin Fazel, Wei Yang, Yulan Liu, Roberto Barra-Chicote, Yixiong Meng, Roland
  Maas, and Jasha Droppo,
\newblock ``{SynthASR: Unlocking Synthetic Data for Speech Recognition},''
\newblock in {\em Proc. Interspeech 2021}, 2021, pp. 896--900.

\bibitem{panayatov2015librispeech}
Vassil Panayotov, Guoguo Chen, Daniel Povey, and Sanjeev Khudanpur,
\newblock ``{Librispeech: An ASR corpus based on public domain audio books},''
\newblock in {\em 2015 IEEE International Conference on Acoustics, Speech and
  Signal Processing (ICASSP)}, 2015, pp. 5206--5210.

\bibitem{kudo2018subword}
Taku Kudo,
\newblock ``Subword regularization: Improving neural network translation models
  with multiple subword candidates,''
\newblock in {\em Proceedings of the 56th Annual Meeting of the Association for
  Computational Linguistics (Volume 1: Long Papers)}, 2018, pp. 66--75.

\bibitem{merity2016pointer}
Stephen Merity, Caiming Xiong, James Bradbury, and Richard Socher,
\newblock ``Pointer sentinel mixture models,''
\newblock in {\em ICLR}, 2017.

\bibitem{johnson2019billion}
Jeff Johnson, Matthijs Douze, and Herv{\'e} J{\'e}gou,
\newblock ``Billion-scale similarity search with {GPUs},''
\newblock {\em IEEE Transactions on Big Data}, vol. 7, no. 3, pp. 535--547,
  2019.

\bibitem{jegou2010product}
Herve Jegou, Matthijs Douze, and Cordelia Schmid,
\newblock ``Product quantization for nearest neighbor search,''
\newblock {\em IEEE transactions on pattern analysis and machine intelligence},
  vol. 33, no. 1, pp. 117--128, 2010.

\end{thebibliography}
